\pdfoutput=1
\documentclass[11pt]{article}

\usepackage[]{EMNLP2022}

\usepackage{times}
\usepackage{latexsym}

\usepackage[T1]{fontenc}
\usepackage[utf8]{inputenc}

\usepackage{microtype}

\usepackage{inconsolata}
\usepackage{latexsym}
\usepackage{times}
\usepackage{helvet}
\usepackage{courier}
\usepackage{amsfonts}
\usepackage{algorithm}
\usepackage{algorithmic}
\usepackage{amssymb}
\usepackage{multirow}
\usepackage{graphicx}
\usepackage{amssymb}
\usepackage{natbib}
\usepackage{amsmath}
\usepackage{appendix}
\usepackage{hyperref}
\title{Fine-grained Category Discovery under Coarse-grained supervision with Hierarchical Weighted Self-contrastive Learning}

\author{
Wenbin An$^{1}$, Feng Tian$^{2}$, Ping Chen$^3$, Siliang Tang$^4$, Qinghua Zheng$^{2}$, QianYing Wang$^{5}$ \\
$^1$ School of Automation Science and Engineering, MOEKLNNS Lab, Xi'an Jiaotong University \\
$^2$ School of Computer Science and Technology, MOEKLNNS Lab, Xi'an Jiaotong University \\
$^3$ Department of Engineering, University of Massachusetts Boston \\
$^4$ College of Computer Science and Technology, Zhejiang University 
$^5$ Lenovo Research\\
\texttt{wenbinan@stu.xjtu.edu.cn, \{fengtian,qhzheng\}@mail.xjtu.edu.cn} \\ 
\texttt{ping.chen@umb.edu, siliang@zju.edu.cn, wangqya@lenovo.com} \\
}

\begin{document}
\maketitle
\begin{abstract}
Novel category discovery aims at adapting models trained on known categories to novel categories. Previous works only focus on the scenario where known and novel categories are of the same granularity.
In this paper, we investigate a new practical scenario called Fine-grained Category Discovery under Coarse-grained supervision (FCDC). FCDC aims at discovering fine-grained categories with only coarse-grained labeled data, which can adapt models to categories of different granularity from known ones and reduce significant labeling cost. 
It is also a challenging task since supervised training on coarse-grained categories tends to focus on inter-class distance (distance between coarse-grained classes) but ignore intra-class distance (distance between fine-grained sub-classes) which is essential for separating fine-grained categories.
Considering most current methods cannot transfer knowledge from coarse-grained level to fine-grained level, we propose a hierarchical weighted self-contrastive network by building a novel weighted self-contrastive module and combining it with supervised learning in a hierarchical manner.
Extensive experiments on public datasets show both effectiveness and efficiency of our model over compared methods.
Code and data are available at \url{https://github.com/Lackel/Hierarchical_Weighted_SCL}.
\end{abstract}

\section{Introduction}
Discovering novel categories based on some known categories has attracted much attention in both Natural Language Processing \citep{thu2021,relation} and Computer Vision \citep{neighbor,dtc}.
Previous works assume that novel categories are of the same granularity (or of the same class hierarchy level) as known categories.
However, in real-world scenarios, novel categories can be more fine-grained sub-categories of known ones (e.g., sports and tennis). A typical application of this scenario is when data analysts want to perform more fine-grained analysis on data with only coarse-grained annotations, where re-labeling fine-grained categories can be time consuming and labour intensive. For example, in the intent detection field, discovering more fine-grained user intents can help to provide better services to customers, but labeling fine-grained intent categories is often much more difficult than labeling coarse-grained ones, since fine-grained annotation often requires higher expertise.
\begin{figure}
\centering
\includegraphics[width=7cm, height=3cm]{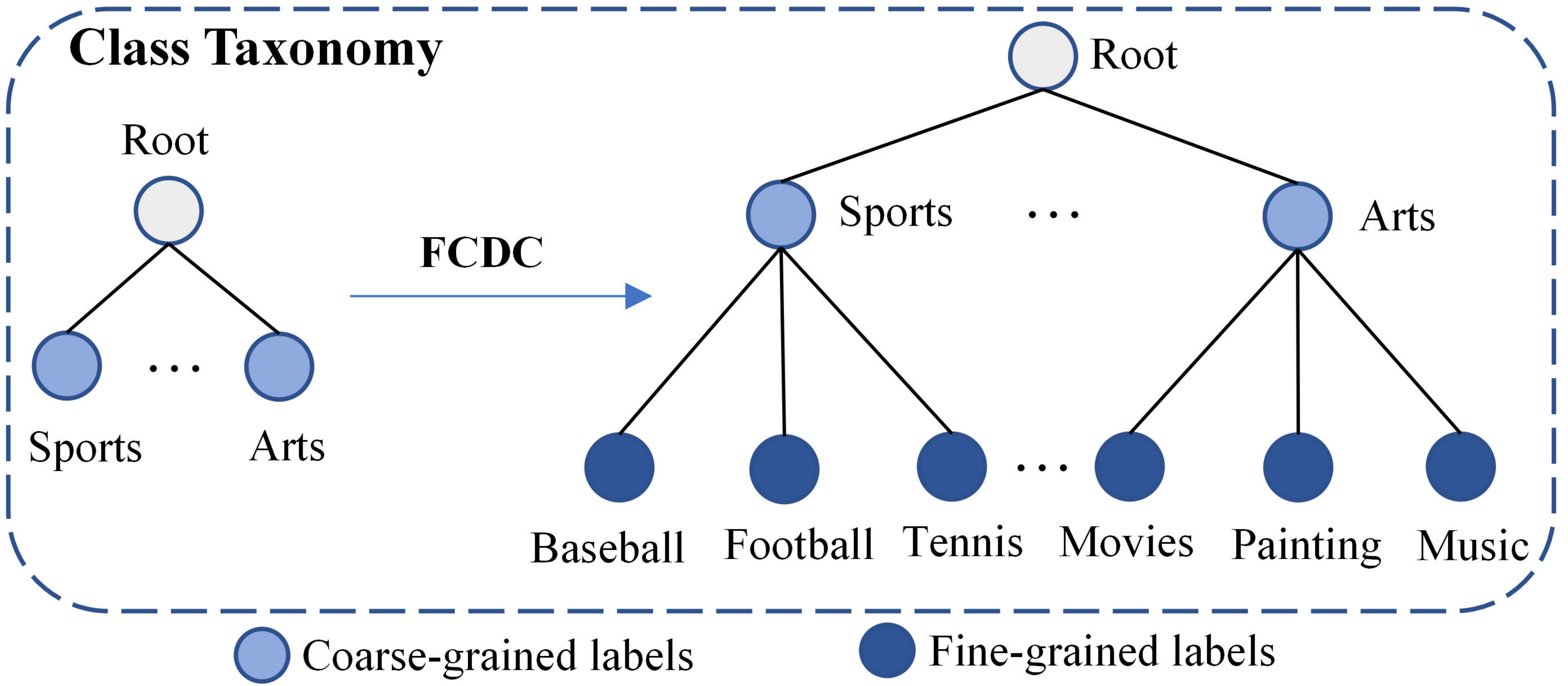}
\caption{An example of proposed FCDC task (fine-grained label names need to be assigned by experts). } 
\label{fig:picture1}
\end{figure}
To meet this requirement, we investigate a new scenario named Fine-grained Category Discovery under Coarse-grained supervision (FCDC). 
As shown in Figure \ref{fig:picture1}, FCDC needs models to discover fine-grained categories (e.g., tennis and music) based only on coarse-grained (e.g., sports and arts) labeled data which are easier and cheaper to obtain. 
% Since performing FCDC only needs coarse-grained labels, most existing classification datasets can be directly used.

In addition to being in line with above practical needs, FCDC is also a challenging task. 
Firstly, performing FCDC requires models to increase intra-class distance to ensure fine-grained separability with only coarse-grained supervision. However, coarse-grained classification only focuses on inter-class distance and does not care about intra-class distance \citep{angular}, so samples with the same coarse-grained labels will be close to each other and hard to be separated in fine-grained feature space. 
Secondly, since fine-grained differentiation depends on correct coarse-grained classification, FCDC also requires models to control inter-class distance to ensure coarse-grained separability. Although increasing intra-class distance can contribute to fine sub-classes separability, it will also decrease inter-class distance, which can result in overlapping between different coarse-grained classes and therefore lead to misclassification. 
So \textbf{how to control and coordinate inter-class and intra-class distance to ensure both coarse-grained and fine-grained separability is the core challenge of FCDC}.

To address above challenges and transfer knowledge from coarse-grained level to fine-grained level, we propose a hierarchical weighted self-contrastive network. By performing different experiments on each layer of BERT, \citet{hierarchy2} find bottom layers of BERT capture more surface features and top layers capture more high-level semantic features, which means BERT can extract features of different granularities from shallow to deep \citep{hierarchy1}. 
Inspired by this phenomenon, the core motivation of our model is to learn coarse-grained knowledge by shallow layers of BERT and learn more fine-grained knowledge by the rest of deep layers hierarchically. This motivation is not only consistent with the feature extraction process of BERT, but also corresponding with the shallow-to-deep learning process of humans.
Specifically, we use given coarse-grained labels to train shallow layers of BERT to learn some surface knowledge, then we propose a weighted self-contrastive module to train deep layers of BERT to learn more fine-grained knowledge based on the learned surface knowledge. 
% By combining supervised learning and contrastive learning in a hierarchical way, our model can fully utilize given coarse-grained knowledge to extract universal features on shallow layers while preserving the ability to extract fine-grained features on deep layers \citep{nature}.

To ensure both coarse-grained and fine-grained separability, we further propose a weighted self-contrastive module to better coordinate inter-class and intra-class distance in the fine-grained feature space.
Specifically, given a query sample, we firstly propose a weighting strategy by weighting different negative samples to control both inter-class and intra-class distance. Then we propose a self-contrastive strategy to generate positive samples to coordinate inter-class and intra-class distance to avoid the overlapping between different coarse-grained classes. We further verify effectiveness and efficiency of our model both theoretically (Section \ref{theory}) and experimentally (Section \ref{main}).

The main contributions of our work can be summarized as threefold:
\begin{itemize}
  \item We propose to investigate a practical scenario called Fine-grained Category Discovery under Coarse-grained supervision (FCDC), we further propose a hierarchical model to learn fine-grained knowledge from shallow to deep to facilitate the FCDC task.
  \item To better coordinate inter-class and intra-class distance, we propose a novel weighted self-contrastive module to ensure both coarse-grained and fine-grained separability.  
  \item Extensive experiments on public datasets show that our model significantly advances best compared methods with a large margin and gets double training efficiency than state-of-the-art contrastive learning methods. 
\end{itemize}

\section{Related work}
\subsection{Contrastive learning}
Contrastive Learning (CL) aims at grouping similar samples closer and separating dissimilar samples far from each other in a self-supervised way \citep{contras2}, which has gained popularity in both Natural Language Processing (NLP) \citep{nlp1} and Computer Vision (CV) \citep{cv1}. A critical point for CL is to build high-quality positive and negative samples. 
The simplest way to construct negative samples is to use other in-batch data as negatives \citep{negative1}. Further, \citet{moco} built a dynamic queue with momentum-updated encoder to keep consistency of representations of negatives. However, these methods consider all negatives equally important, which may lose discriminative information of different negatives.
As for positive samples, in CV, one common way is taking two different transformations of the same image as the query and positive sample \citep{positivecv}. And in NLP, augmentation techniques such as word deletion \citep{nlp2}, adversarial attack \citep{adv} and dropout \citep{simcse} were proposed to generate positives. 
Although there are some recent works \citep{self} using outputs from different levels of a network as positives, we have totally different motivations: they aim at providing more high-quality positives for representation learning while we aim at better adjusting intra-class and inter-class distance.

% To mitigate this issue, we give different negatives with different weight to better regulate distance between queries and negatives.

% To mitigate this issue, we propose a self-contrastive module by taking shallow representations of queries extracted from bottom layers of BERT as positives. Since we only need to perform forward and back propagation once, our training efficiency is double than above methods. 

\begin{figure*}
\centering
\includegraphics[width=13cm, height=7cm]{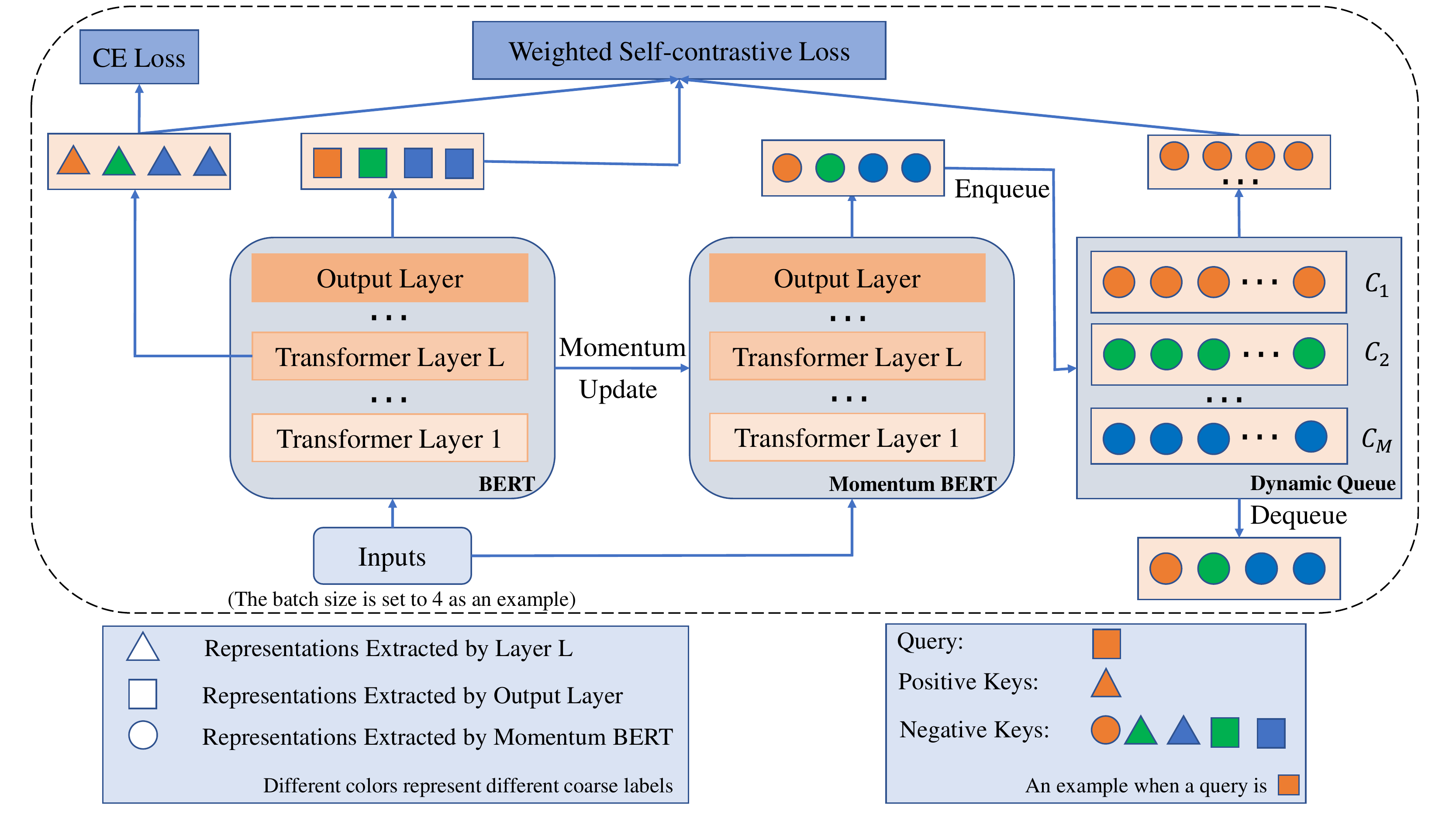}
\caption{The overall architecture of our model. CE means Cross Entropy.} 
\label{fig:picture7}
\end{figure*}

\subsection{Novel Category Discovery}
With data volume increases, novel categories especially novel fine-grained categories may be introduced into datasets \citep{coarse2fine}. To discover novel categories without human annotation, most previous work adopted clustering methods and transfer learning methods to generate pseudo labels for unlabeled data to train their models \citep{online}. For example, \citet{thu2021} proposed an alignment strategy to perform DeepCluster \citep{deepcluster} to discover novel categories. \citet{mutual} proposed a mutual mean teaching network to refine noisy pseudo labels to perform unsupervised person re-identification. 
Recently, Two similar tasks as FCDC are proposed. \citet{angular} proposed to perform fine-grained image classification under coarse-grained supervision with angular contrastive learning, and they performed this task in a few-shot learning way which needs extra fine-grained labels for each categories. \citet{coarse2fine} proposed to perform fine-grained text classification with coarse-grained annotations, but they need extra fine-grained label hierarchy and corresponding label names to assist in the task. These two tasks both rely on extra fine-grained knowledge from human annotations, which is usually unavailable when novel categories appear in real-world applications. Comparatively, our FCDC is a category discovery task which does not require fine-grained knowledge and is more adapted to real world scenarios.

\subsection{Problem Formulation}
\label{problem}
% The proposed FCDC task has two objectives: discovering fine-grained classes from scratch and classifying inputs into proper fine-grained categories.
Denote by $\mathcal{Y}_{coarse}=\{\mathcal{C}_{1}, \mathcal{C}_{2},...,\mathcal{C}_{M}\}$ a set of coarse-grained classes.
The training set of FCDC is a set of texts $\mathcal{D}_{train}=\{\mathcal{D}_{1},\mathcal{D}_{2},...,\mathcal{D}_{N}\}$ with their coarse-grained labels $\{\textit{c}_{1},\textit{c}_{2},...,\textit{c}_{N}\}$, where $\textit{c}_{i} \in \mathcal{Y}_{coarse}$.
Different from previous tasks \citep{angular,coarse2fine} where the fine-grained label set $\mathcal{Y}_{fine} = \{\mathcal{F}_{1}, \mathcal{F}_{2},...,\mathcal{F}_{K}\}$ is already known, FCDC assumes that we do not have any prior knowledge about fine-grained labels.
So FCDC requires models to perform clustering methods (e.g., K-Means) to discover fine-grained clusters $\mathcal{Y}_{fine}$ with $\mathcal{D}_{train}$. Since performing clustering will assign each input with a specific cluster assignment, FCDC can also classify inputs into proper fine-grained categories $\{\textit{f}_{1},\textit{f}_{2},...,\textit{f}_{N}\}$.
Although the number of fine-grained clusters $\textit{K}$ can be estimated with various methods from the clustering area, we assume it is known in FCDC following previous similar works \citep{thu2021,angular} to make a fair comparison.

\section{Proposed Approach}
As shown in Figure \ref{fig:picture7}, our model mainly contains three components: BERT, Dynamic Queue and Momentum BERT. BERT is used to extract both coarse-grained and fine-grained features. Dynamic Queue can store more negative samples grouping by their coarse-grained labels following  \citet{angular}. Momentum BERT is used to update representations of samples in Dynamic Queue. 
Inspired by the "shallow to deep" learning process of humankind and the ability of pre-trained models to extract features from coarse-grained to fine-grained \citep{hierarchy2,hierarchy1}, a core motivation of our model is to learn fine-grained knowledge in a progressive way. Specifically, our model can learn coarse-grained knowledge at shallow layers under coarse-grained supervision and learn more fine-grained knowledge at deep layers with the proposed weighted self-contrastive learning.

\subsection{Supervised Learning}
We firstly perform supervised learning on Transformer layer \textit{L} of BERT to learn coarse-grained knowledge. Given the \textit{i-th} document $\mathcal{D}_{i}$ with its coarse-grained label $\textit{c}_{i}$, we use all token embeddings from the \textit{L}-th layer of BERT as its shallow features. Then we apply a mean-pooling layer to get its shallow feature representation $h_{i}^{L}$:
\begin{equation}
    h_{i}^{L} = mean\raisebox{0mm}{-}pooling(BERT_{L}(\mathcal{D}_{i}))
\end{equation}

where $h_{i}^{L} \in \mathbb{R}^{h}$ is the hidden state of feature representations, $h$ is the dimension of hidden representations. Then we perform supervised learning with cross entropy loss on coarse-grained labels to get supervised loss $\mathcal{L}_{sup}^{L}$ at layer \textit{L}:
\begin{gather}
    z_{i}^{L} = \sigma (W_{a}h_{i}^{L} + b_{a}) \\
    \mathcal{L}_{sup}^{L} = \mathbb{-} \frac{1}{N} \sum_{i=1}^{N} log \frac{exp(({z}_{i}^{L})^{c_{i}})}{\sum_{j=1}^{K}exp((z_{i}^{L})^{j})}
\end{gather}
where $z_{i}^{L} \in \mathbb{R}^{M}$ is the output logits, $M$ is the number of coarse-grained classes. $\sigma$ is the Tanh activation function, $W_{a} \in \mathbb{R}^{h*M}$ and $b_{a} \in \mathbb{R}^{M}$ are learnable weights and bias terms, respectively.  $(z_{i})^{j}$ is the \textit{j}-th element of output logits $z_{i}$.

% The advantages of performing supervised learning in layer \textit{L} lies in two aspects: providing basis for learning fine-grained knowledge and mitigating overfitting problem. Firstly, by forcing the model to learn coarse-grained knowledge in shallow layers, our model can learn more fine-grained knowledge in deeper layers based on learned coarse-grained knowledge in a progressive way. Secondly, compared with adding coarse-grained supervision on the output layer and training on the entire model, our strategy can avoid the model overfitting to coarse-grained labels by reducing model complexity (reducing from 12 transformer layers to \textit{L} transformer layers).

\subsection{Weighted Self-contrastive Learning}
\label{goal}
\begin{figure}
\centering
\includegraphics[width=7.5cm, height=2.2cm]{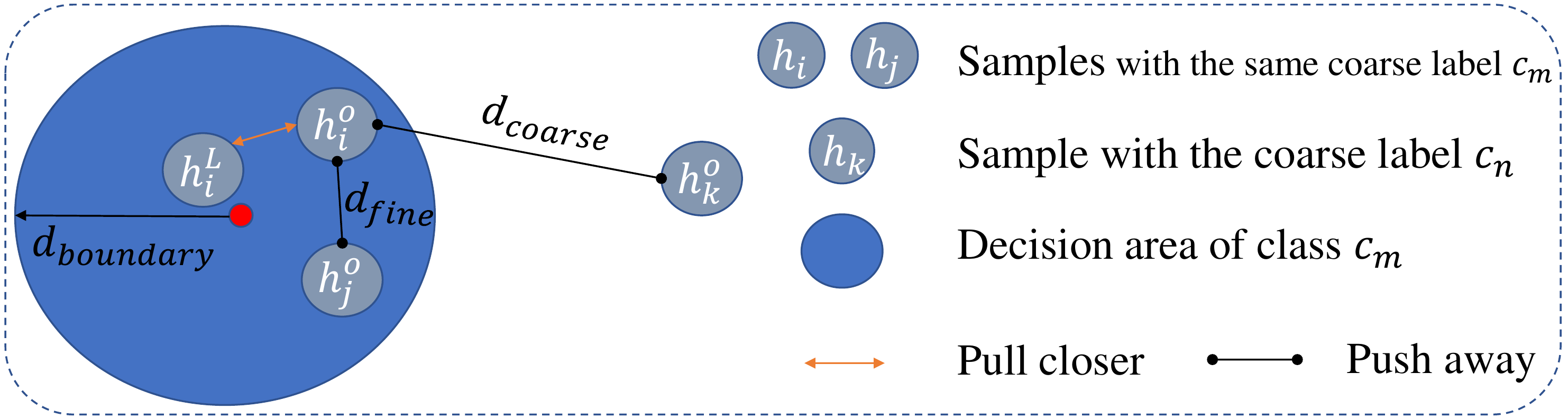}
\caption{The effectiveness of our self-contrastive module, which can ensure both intra-class and inter-class distance.} 
\label{fig:picture4}
\end{figure}

As shown in Figure \ref{fig:picture4}, denote the coarse-grained inter-class and intra-class distance by $d_{coarse}$ and $d_{fine}$, respectively.
Supervised learning on coarse-grained labels can ensure $d_{coarse} \gg 0$ but will also make $d_{fine} \approx 0$, which can bring difficulties for fine-grained categorization. So how to increase $d_{fine}$ to ensure separability of fine-grained sub-classes is a severe challenge. Meanwhile, increasing $d_{fine}$ without restraint will result in overlapping between different coarse-grained classes and therefore lead to misclassification. So how to constrain $d_{fine}$ to ensure the proper classification on coarse-grained classes is the other challenge. In summary, our total goal can be described as: 
\begin{equation}
\label{equation1}
    0 \ll d_{fine} < d_{boundary} \ll d_{coarse}
\end{equation}
where $d_{boundary}$ is a threshold to ensure that samples fall into proper coarse-grained classes.

To achieve above objectives, we propose a weighted self-contrastive module by introducing a novel generation strategy for positive samples and a weighting strategy for negative samples.

\subsubsection{Negative Key Generation}
Given the \textit{i-th} document $\mathcal{D}_{i}$, we use all token embeddings from the output layer of BERT as its deep features. Then we apply a mean-pooling layer to get its deep feature representation $h_{i}^{o} \in \mathbb{R}^{h}$:
\begin{equation}
    h_{i}^{o} = mean\raisebox{0mm}{-}pooling(BERT_{o}(\mathcal{D}_{i}))
\end{equation}

\noindent \textbf{In-batch negative keys}\quad  Given $h_{i}^{o}$ with its coarse-grained label $c_{i}$ as a query $q$, we treat shallow and deep features of other in-batch samples as its in-batch negative keys, where $k_{-}^{in}(i) = \{h^{L}_{j}, h^{o}_{j} \}_{j=1...N, j \neq i}$. In this way, we can increase distance between different samples so that satisfying $ d_{fine} \gg 0$ and $d_{coarse} \gg 0$. To satisfy $d_{coarse} \gg d_{fine}$, we propose a weighting strategy by giving more weights to samples with different coarse-grained labels as the query $q$ to further increase their distance. So $k_{-}^{in}$ can be divided into two groups according to the coarse-grained labels:
\begin{gather}
    k_{-}^{diff}(i) = \{k \in k_{-}^{in}(i)  : c_{k} \neq c_{i}\} \\
    k_{-}^{same}(i) = \{k \in k_{-}^{in}(i)  : c_{k} = c_{i}\}
\end{gather}

\noindent \textbf{Momentum negative keys}\quad To provide more negative keys, we build a momentum BERT and a set of dynamic queues $\{\mathcal{Q}_{i}\}_{i=1}^{M}$ to store previous samples grouped by their coarse-grained labels following \citet{angular}, where \textit{M} is the number of coarse-grained classes. Specifically, given $h_{i}^{o}$ with its coarse-grained label $c_{i}$ as a query, we treat samples from the queue $\mathcal{Q}_{c_{i}}$ as its momentum negative keys:
\begin{equation}
    k_{-}^{m}(i) =  \{k \in \mathcal{Q}_{c_{i}} \}
\end{equation}
Feature representations of samples in dynamic queues are extracted by momentum BERT, and the parameters of momentum BERT are updated in a momentum way \citep{moco}. At the end of each iteration, the dynamic queues will be updated by adding novel samples and removing the earliest samples. Since samples in $k_{-}^{m}(i)$ have the same coarse-grained label as the query, they are much harder to be separated and beneficial to better representation learning. 

The overall negative keys for the query $h_{i}^{o}$ is :
\begin{equation}
    k_{-}(i) = \{k_{-}^{diff}(i), k_{-}^{same}(i), k_{-}^{m}(i)\}
\end{equation}

\subsubsection{Positive Key Generation}
\label{positive}
By weighting different negative samples, we can satisfy the condition $0 \ll d_{fine} \ll d_{coarse}$. But increasing $d_{fine}$ without restraint will violate the condition $ d_{fine} < d_{boundary}$ and make some samples fall into incorrect coarse-grained classes. To solve this problem, we propose a self-contrastive strategy by treating shallow features of a query as its positive key. Specifically, given the deep feature representation $h_{i}^{o}$ for document $\mathcal{D}_{i}$ as a query, we treat $h_{i}^{L}$ as its positive key:
\begin{equation}
    k_{+}(i) = h_{i}^{L}
\end{equation}
As shown in Figure \ref{fig:picture4}, after supervised learning on coarse-grained labels at layer \textit{L}, $h_{i}^{L}$ can be very close to the class center of $c_{i}$, so pulling $h_{i}^{o}$ close to $h_{i}^{L}$ will also pull $h_{i}^{o}$ close to the class center of $c_{i}$. In this way, we can increase $d_{fine}$ with restraint and satisfy the condition $ d_{fine} < d_{boundary}$ without computing the specific value of $d_{boundary}$. Another advantage of our self-contrastive strategy is that we can get double training efficiency than traditional data augmentation-based methods \citep{nlp1,simcse} since we only need to perform forward and backward propagation only once to get and update both queries and positive keys (Section \ref{efficiency}).

\subsubsection{Weighted Self-contrastive Loss}
Given the query $h_{i}^{o}$ with its positive key $k_{+}(i)$ and negative keys $k_{-}(i)$, the overall loss of our weighted self-contrastive module is:
\begin{equation}
     \mathcal{L}_{cont} = \sum_{i=1}^{N} \mathbb{-} log  \frac{e^{sim(h_{i}^{o}, h_{i}^{L})/\tau}}{\sum\limits_{l \in k_{-}(i)}\!\alpha_{l}\!\sum\limits_{k \in l}e^{sim(h_{i}^{o}, h_{k})/\tau}}
\end{equation}
where $\alpha_{l} \in \{ \alpha_{same}, \alpha_{diff}, \alpha_{m} \}$ are weighting factors for different negative keys, $sim(h_{i}, h_{j})$ is cosine similarity $\frac{h_{i}^{T}h_{j}}{\left\|h_{i}\right\| \cdot \left\|h_{j}\right\|}$ and $\tau$ is a temperature hyperparameter.

By weighting different negative keys and selecting shallow features as the positive key, our model can satisfy the goal in Inequation \ref{equation1} and provide conditions for subsequent fine-grained categorization.

\subsubsection{Theoretical Analysis}
\label{theory}
The effectiveness of our weighted self-contrastive learning compared with traditional contrastive learning from the gradient perspective is analyzed below.

\noindent \textbf{Self-contrastive Strategy}\quad Compared with traditional contrastive loss which only aims at grouping queries and their transformations closer, our self-contrastive strategy aims at pulling queries and their shallow features closer:

\begin{equation}
\begin{aligned}
     sim(h_{i}^{o}, h_{i}^{L}) := sim(h_{i}^{o}, h_{i}^{L}) + \frac{1}{\tau}
\end{aligned}
\end{equation}

Since $\tau$ is positive, the positive similarity will increase and $h_{i}^{o}$ will be grouped closer to $h_{i}^{L}$. After supervised learning on coarse-grained labels at layer \textit{L}, $h_{i}^{L}$ can be close to the class center of $c_{i}$, so pulling $h_{i}^{o}$ closer to $h_{i}^{L}$ will also pull $h_{i}^{o}$ closer to the class center of $c_{i}$. 
So our Self-Contrastive strategy can guarantee queries fall into correct coarse-grained categories and get double training efficiency since we only need to perform forward and backward propagation only once to get and update both queries and positive keys.

% Since negatives with the same coarse-grained labels as queries have larger gradients (Appendix \ref{gradient}), traditional contrastive loss will push these negatives farther from queries than those with different coarse-grained labels as queries, which leads to $d_{coarse} < d_{fine}$ and is opposite of what we expect to solve the FCDC task.

\noindent \textbf{Weighting Strategy}\quad Since negatives with the same coarse-grained labels as queries have larger gradients \citep{understand}, traditional contrastive loss will push these negatives farther from queries than those with different coarse-grained labels as queries, which leads to $d_{coarse} < d_{fine}$ and is opposite of what we expect to solve the FCDC task. To mitigate this limitation, we propose a weighting strategy to give more weights to samples with different coarse-grained labels as the query to further increase their distance:

\begin{equation}
\begin{aligned}
     sim(h_{i}^{o}, h_{j}^{o}) := sim(h_{i}^{o}, h_{j}^{o}) - \alpha_{l} \cdot \mathcal{P}_{i,j}
\end{aligned}
\end{equation}

\begin{equation}
    \mathcal{P}_{i,j} = \frac{1}{\tau} \cdot \frac{e^{sim(h_{i}^{o}, h_{j}^{o})/\tau}}{{\sum\limits_{l \in k_{-}(i)}\!\alpha_{l}\!\sum\limits_{k \in l}e^{sim(h_{i}^{o}, h_{k})/\tau}}}
\end{equation}

By increasing the weighting factor $\alpha_{l}$ for negatives with different coarse-grained labels as queries, the corresponding similarity will decrease faster. So negatives with different coarse-grained labels from queries will be pushed farther than those with the same coarse-grained labels as queries, which can guarantee $d_{fine} < d_{coarse}$ for the FCDC task.

\begin{table}[t]
\centering
\begin{tabular}{lccccc}
\hline
Dataset & |$\mathcal{C}$| & |$\mathcal{F}$| & \# Train  & \# Test  \\
\hline
CLINC   &  10    &  150       & 18,000     &  1,000    \\
WOS     &  7     &  33        & 8,362      &  2,420     \\
HWU64   &  18    &  64        & 8,954      &  1,031     \\
\hline
\end{tabular}
\caption{Statistics of datasets. \# indicates the number of samples. |$\mathcal{C}$|, |$\mathcal{F}$| means the number of coarse-grained and fine-grained classes, respectively.}
\label{table1}
\end{table}

\begin{table*}
\centering
\begin{tabular}{l|ccc|ccc|ccc}
\hline
\multirow{2}*{Methods} & \multicolumn{3}{c|}{CLINC} &\multicolumn{3}{c|}{WOS} &\multicolumn{3}{c}{HWU64}\\
\cline{2-10}
                           &ACC    &ARI    &NMI    &ACC    &ARI    &NMI     &ACC    &ARI    &NMI\\
\hline
Unsupervised               &33.38  &16.42  &63.46  &32.32  &18.21  &47.12   &33.66  &16.88  &56.78\\             
Coarse Supervised          &45.91  &32.27  &75.04  &39.42  &33.67  &61.60   &42.41  &33.74  &71.57\\                               
\hline
DeepCluster                &26.40  &12.51  &61.26  &29.17  &18.05  &43.34   &29.74  &13.98  &53.27\\
DeepAligned                &29.16  &14.15  &62.78  &28.47  &15.94  &43.52   &29.14  &12.89  &52.99\\
SimCSE                     &40.22  &23.57  &69.02  &25.87  &13.03  &38.53   &24.48  &8.42  &46.94\\
Anchor                     &45.60  &33.11  &75.23  &41.20  &37.00  &65.42   &37.34  &34.75  &74.99\\
Delete One Word            &47.11  &31.28  &73.39  &24.50  &11.68  &35.47   &21.30  &6.52  &44.13\\
\hline
DeepCluster + CE           &30.28  &13.56  &62.38  &38.76  &35.21  &60.30   &41.73  &27.81  &66.81\\
DeepAligned + CE           &42.09  &28.09  &72.78  &39.42  &33.67  &61.60   &42.19  &28.15  &66.50\\
Anchor + CE                &44.44  &31.50  &74.67  &39.34  &26.14  &54.35   &32.90  &30.71  &74.73\\
Delete One Word + CE       &47.87  &33.79  &76.25  &41.53  &33.78  &61.01   &35.13  &31.84  &74.88\\
SimCSE + CE                &52.53  &37.03  &77.39  &41.28  &34.47  &61.62   &34.04  &31.81  &74.86\\
Ours                       &\textbf{74.67}  &\textbf{64.77}  &\textbf{89.14}             &\textbf{63.64}           &\textbf{51.55}  &\textbf{72.46}   &\textbf{58.45}           &\textbf{48.20}  &\textbf{78.66}\\
% Ours on coarse             &\textbf{98.71}  &\textbf{97.16}  &\textbf{96.70}             &\textbf{68.04}           &\textbf{56.16}  &\textbf{73.69}  \\
\hline
\end{tabular}
\caption{Model comparison results (\%) on fine-grained categories. Average results over 5 runs are reported. '+ CE' means adding coarse-grained supervision with cross entropy loss. We also perform statistical significance test and all the p-values are less than $10^{-6}$, which means our improvement is significant.}
\label{table2}
\end{table*}

\subsection{Overall Loss}
To further guarantee samples to be classified into proper coarse-grained categories, 
we also add supervised learning on coarse-grained labels at the output layer. So the overall loss for our hierarchical weighted self-contrastive network is:
\begin{equation}
     \mathcal{L} = \mathcal{L}_{sup}^{o} + \gamma_{1} \mathcal{L}_{sup}^{L} + \gamma_{2} \mathcal{L}_{cont}
\end{equation}
where $\mathcal{L}_{sup}^{o}$ is the cross entropy loss at the output layer. $\gamma_{1}$ and $\gamma_{2}$ are weighting factors.

After representation learning, we simply perform the non-parametric clustering method K-Means to discover fine-grained categories based on features extracted by the output layer of BERT.

\section{Experiments}
\subsection{Datasets}
To evaluate effectiveness of our model, we conduct experiments on three public datasets. Statistics of three datasets can be found in Table \ref{table1}.

\noindent \textbf{CLINC} is an intent classification dataset released by \citet{clinc}.

\noindent \textbf{Web of Science (WOS)} is a paper classification dataset released by \citet{wos}.

\noindent \textbf{HWU64} is a personal assistant query classification dataset released by \citet{hwu64}.

\subsection{Implementation Details}
\label{details}
We use the pre-trained BERT model (bert-base-uncased) implemented by Pytorch \citep{transformers} as our backbone and adopt most of its suggested hyper-parameters. We use the cuml library \citep{machine} to perform K-Means on GPU to speed up calculations. We use the AdamW optimizer with 0.01 weight decay. Gradient clipping is also used with the norm 1.0.
For hyper-parameters, temperature $\tau$ is set to 0.1, layer \textit{L} is set to 8, and the weighting factors $\alpha_{l}$ for $\{k_{-}^{diff}(i), k_{-}^{same}(i), k_{-}^{m}(i)\}$ are set to \{1.4, 1.0, 1.0\}, weighting factors $\{\gamma_{1}, \gamma_{2}\}$ are set to \{0.001, 0.008\}.
The training batch size is set to 128, and the testing batch size is set to 64. The momentum queue size for each coarse-grained category is set to 128, and the momentum factor for Momentum BERT is set to 0.9. The hidden dimension $h$ is 768, the learning rate is set to $5e^{-5}$, the dropout rate is set to 0.1. The training epoch is set to 20.

For a fair comparison, we use the same BERT model as ours to extract features for all compared methods and adopt hyper-parameters in their original papers.

\subsection{Compared Methods}

\noindent \textbf{Baselines}\quad We perform FCDC with BERT in unsupervised and coarse-supervised way as baselines. 

\noindent \textbf{Self-supervised Methods}\quad DeepCluster \citep{deepcluster} and DeepAligned \citep{thu2021} are self-supervised methods using self-training techniques and achieve state-of-the-art results in many category discovery tasks. Ancor \citep{angular} is a self-supervised method designed for few-shot fine-grained classification with coarse-grained labels. SimCSE \citep{simcse} and Delete One Word \citep{nlp1} are contrastive learning methods in NLP with different data augmentation techniques.

\noindent \textbf{Self-supervised + Cross Entropy}\quad  To investigate the influence of coarse-grained supervision on compared models, we further add cross entropy loss on coarse-grained labels $\mathcal{L}_{sup}^{o}$ to their loss function.

\subsection{Evaluation Metrics}
We use fine-grained labels as ground truth to evaluate model performance on testing sets.
Since no fine-grained knowledge is available for the FCDC task, we need to perform clustering to discover fine-grained categories. 
Clustering performance can reflect the quality of discovered fine-grained clusters (More compact clusters usually mean better discovered categories). And classification performance can reflect the semantic overlap between discovered clusters and real categories.

To evaluate clustering performance, we use two broadly used external evaluation metrics. Adjusted Rand Index (ARI) is used to evaluate the degree of agreement between cluster assignments and ground truth. And Normalized Mutual Information (NMI) is used to evaluate the mutual information between cluster assignments and ground truth.
To evaluate classification performance, we use metric Accuracy (ACC), which is obtained from Hungarian algorithm \citep{hungarian} to align cluster assignments and ground truth.

\subsection{Main Results}
\label{main}
\textbf{Model performance on fine-grained categories} are reported in Table \ref{table2}. From the results we can draw following conclusions. 
Our model significantly outperforms other compared methods across all datasets. We contribute reasons of better performance of our model to following two points. Firstly, we propose a hierarchical architecture to learn fine-grained knowledge from shallow to deep, which is consistent with the feature extraction process of BERT and the shallow-to-deep learning process of humans. Secondly, we propose a weighted self-contrastive module to coordinate inter-class and intra-class distance so that we can better learn both coarse-grained and fine-grained knowledge. 
Self-training methods perform badly on all datasets and evaluation metrics since they rely on abundant labeled data to generate high-quality pseudo labels for unlabeled data. Contrastive learning methods perform better than self-training methods since they do not need fine-grained labels to initialize their models. However, their performance is still much worse than ours since they cannot fully utilize given coarse-grained labels to control inter-class and intra-class distance between samples. 
We also find that model performance of most compared methods increases with the addition of coarse-grained supervision, which means coarse-grained supervision can boost model performance on fine-grained tasks.

\textbf{Our model performance on coarse-grained categories} are reported in Table \ref{table20}. From the table we can see that our model gets similar classification accuracy to the upper-bound coarse-supervised BERT, which means that our model can control not only intra-class distance to ensure fine-grained  separability, but also inter-class distance to ensure coarse-grained variability.

\begin{table}
\setlength\tabcolsep{8pt}
\caption{Classification accuracy (\%) on coarse-grained categories on test sets.}
\centering
\begin{tabular}{l|c|c}
\hline
\textbf{Model} & CLINC & WOS \\
\hline
Coarse Supervised      & 98.58 & 91.86 \\
Ours                   & 98.71 & 91.45 \\
\hline
\end{tabular}
\label{table20}
\end{table}

\begin{table}
\setlength\tabcolsep{8pt}
\caption{Results (\%) of different model variants. '-' means that we remove the component from our model.}
\centering
\begin{tabular}{l|c|c|c}
\hline
\textbf{Model} & ACC & ARI & NMI \\
\hline
ALL & 74.67 & 64.77 & 89.14\\
- Momentum                     & 73.38 & 64.47 & 88.91\\
- $\mathcal{L}_{sup}^{L}$      & 72.93 & 63.72 & 88.06\\
- Weighting                    & 71.75 & 62.99 & 88.47\\
- Self-Contrast                & 53.21 & 40.05 & 75.36\\
\hline
\end{tabular}
\label{table3}
\end{table}

\section{Discussion}
\subsection{Ablation Study}
\label{ablation}
To investigate contributions of different components to our model, we compare the performance of our model with its variants on the CLINC dataset. As shown in Table \ref{table3}, removing different components will affect model performance more or less, which indicates the effectiveness of different components of our model.
Removing Momentum Encoder has minimal impact, since our model is insensitive to the number of negative samples. 
Removing weighting strategy or cross entropy loss at shallow layers also hurt model performance since they can help to learn coarse-grained knowledge and lay foundation for learning fine-grained knowledge. Above all, removing self-contrastive strategy results in a significant decrease, since it is responsible for controlling intra-class and inter-class distance.

\begin{figure}
\centering
\includegraphics[width=7.5cm, height=5cm]{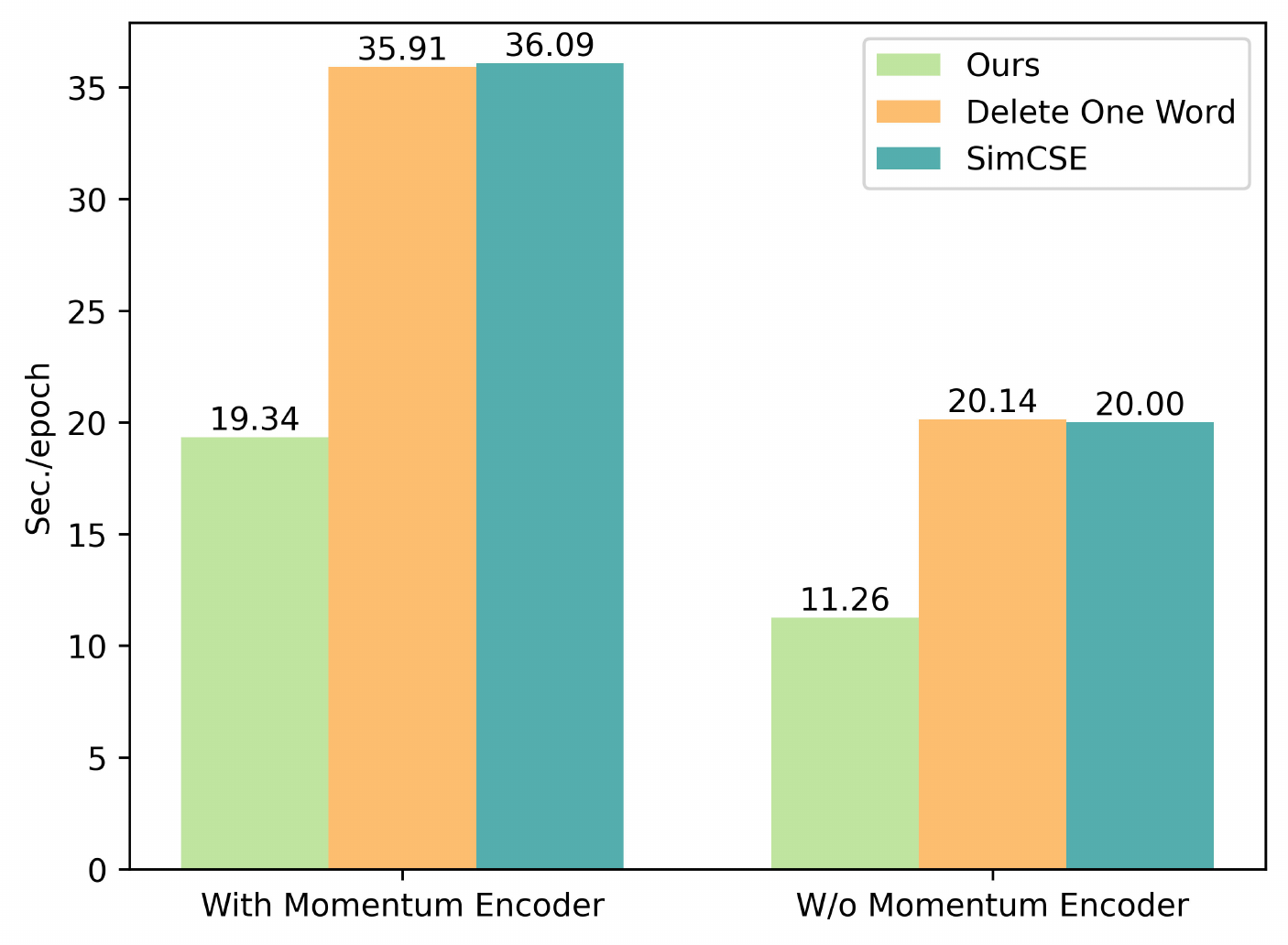}
\caption{Training efficiency comparison. } 
\label{fig:picture6}
\end{figure}

\subsection{Training Efficiency}
\label{efficiency}
In this section, we compare the training efficiency of our model with contrastive methods SimCSE and Delete One Word on the CLINC dataset. 
We test all methods using the BERT base model trained on the same hardware platform (an AMD EPYC CPU 7702 and a RTX 3090 GPU) with the batch size 128. Average results over 100 epochs are shown in Figure \ref{fig:picture6}. Compared with SimCSE and Delete One Word, our model gets double training efficiency both when adding or removing Momentum Encoder, which benefits from our self-contrastive strategy.
Specifically, our model utilizes shallow features of queries as positive keys, which only needs to perform forward and backward propagation once to get and update both queries and positive keys.

% \subsection{Visualization}
% We visualize the learned embeddings of our model on the CLINC dataset using t-SNE \citep{tsne} in Figure \ref{fig:picture5}. It can be seen that our model can ensure both inter-class and intra-class distance to facilitate the FCDC task. 
% Specifically, our model can separate different coarse-grained categories with a large margin benefiting from the supervised learning on coarse-grained labels. Meanwhile, different from traditional supervised learning methods which usually ignore the intra-class distance, our model can better increase the distance of samples within the same coarse-grained categories to ensure the intra-class separability, which benefits from the proposed weighted self-contrastive module.

\subsection{Visualization}
\begin{figure}
\centering
\includegraphics[width=7.5cm, height=6cm]{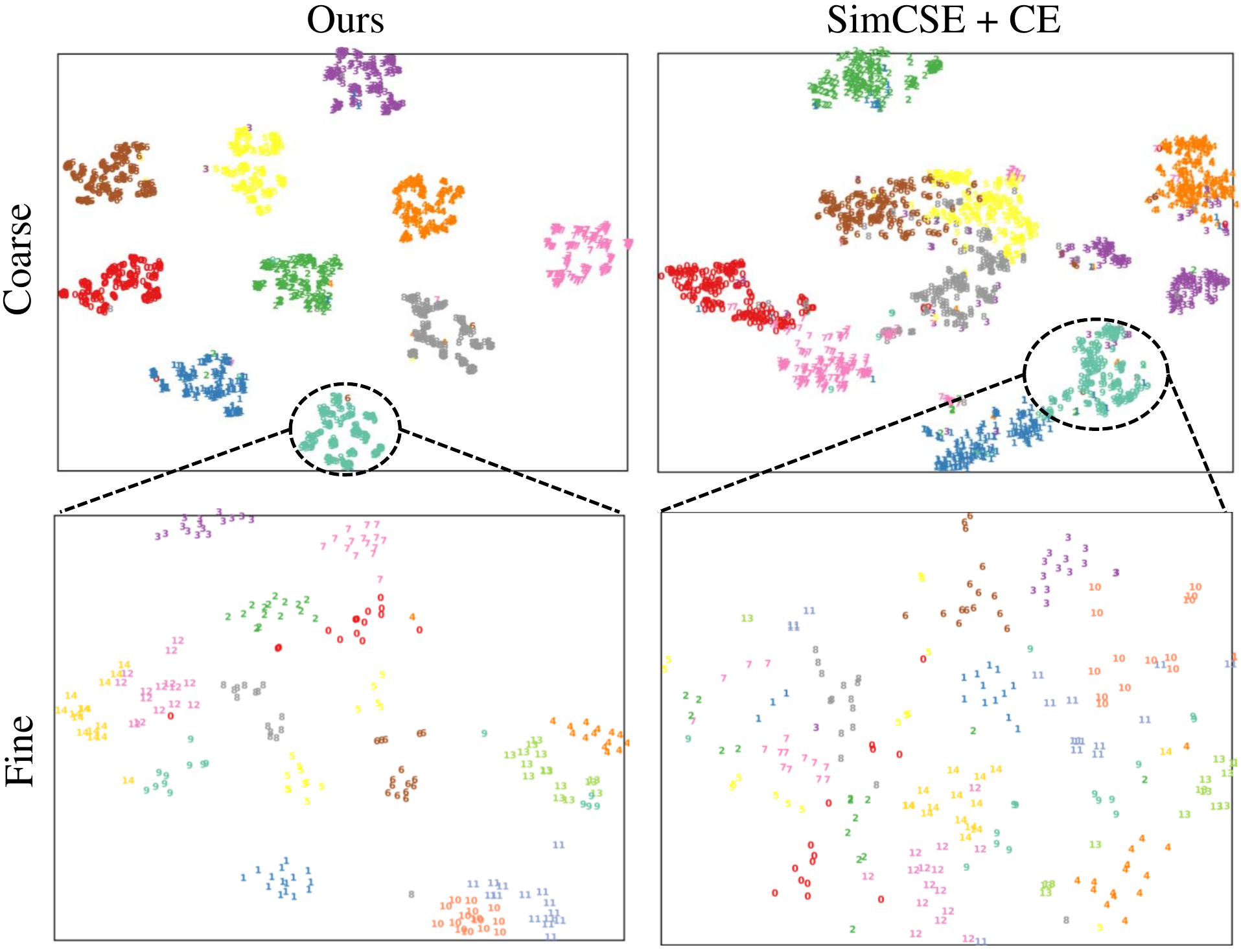}
\caption{TSNE visualization of learned embeddings. \textbf{Top}: coarse-grained categories. \textbf{Bottom}: fine-grained categories of one arbitrary coarse-grained category. \textbf{Left}: Ours. \textbf{Right}: SimCSE + CE.} 
\label{fig:picture10}
\end{figure}

We further visualize the learned embeddings of our model and SimCSE using t-SNE on the CLINC dataset in Figure \ref{fig:picture10}. 
Our model can separate different coarse-grained categories with a larger margin than SimCSE (Top in Figure \ref{fig:picture10}), which benefits from our strategy of combining supervised learning and contrastive learning in a hierarchical way. 
Furthermore, our model can also separate different fine-grained categories with a larger margin (Bottom in Figure \ref{fig:picture10}), which benefits from the weighted self-contrastive module. 
In summary, our model can better control both inter-class and intra-class distance between samples to facilitate the FCDC task than traditional contrastive learning methods.

\subsection{Choices of \textit{L} and weighting factors}
\begin{figure}
\centering
\includegraphics[width=7cm, height=4.5cm]{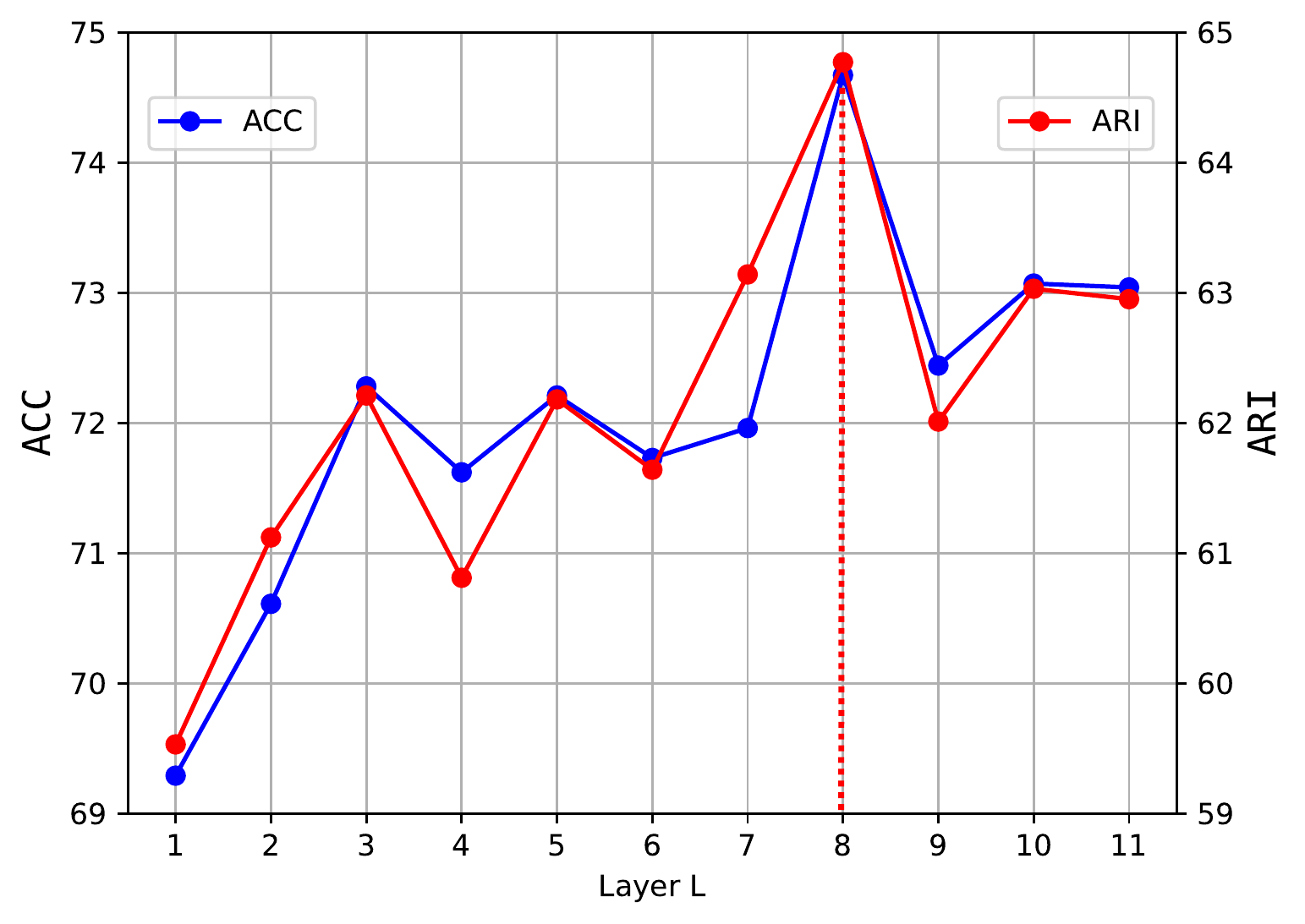}
\caption{Effect of shallow layer \textit{L}.} 
\label{fig:picture5}
\end{figure}

\noindent \textbf{Effect of Shallow Layer \textit{L}}\quad The influence of the choice of shallow layer \textit{L} on model performance is shown in Figure \ref{fig:picture5}. Our model achieves the best performance when \textit{L}=8. In this way, our model can learn coarse-grained knowledge at shallow layers (\textit{L}<8) and provide enough model capacity to learn fine-grained knowledge at deeper layers (\textit{L}>8), which is consistent with the feature extraction process of BERT \citep{hierarchy2}.

\begin{figure}
\centering
\includegraphics[width=7cm, height=4.5cm]{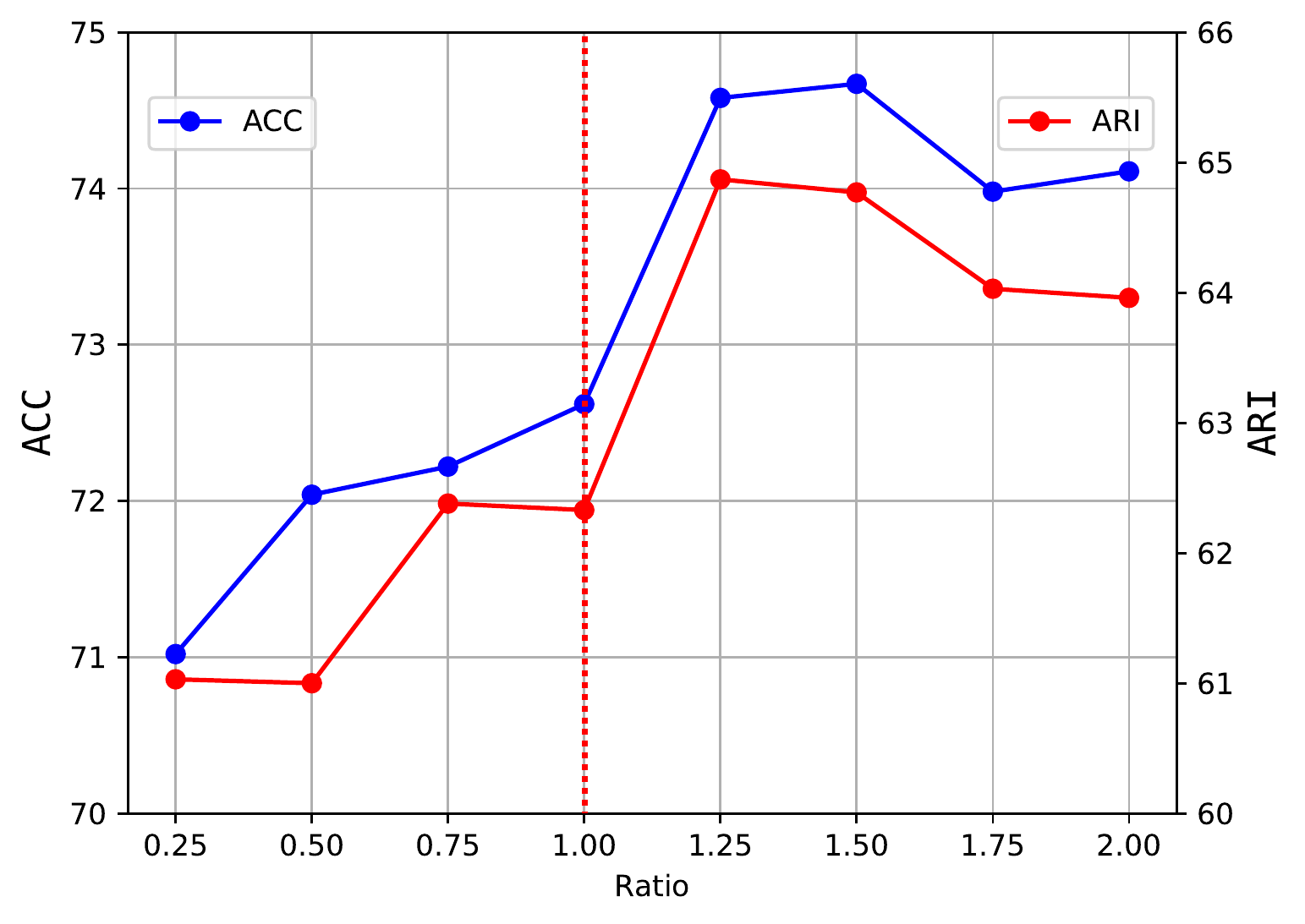}
\caption{Effect of ratio $\beta = \alpha_{diff} / \alpha_{same}$.} 
\label{fig:picture111}
\end{figure}

% The model capacity that needs to be trained by contrastive learning will increase with \textit{L} decreasing. And 
\noindent \textbf{Effect of Weighting Factors}\quad We investigate the influence of the ratio $\beta = \alpha_{diff} / \alpha_{same}$ in Figure \ref{fig:picture111} (We fixed $\alpha_{m}=1$ since it has little influence). As analyzed in Section \ref{theory}, by giving more weights to negatives with different coarse-grained labels as queries ($\beta > 1$), our weighting strategy can keep these negatives further away from queries and guarantee $d_{fine} < d_{coarse}$. On the contrary, when $\beta < 1$, negatives with the same coarse-grained labels as queries will be further away from the queries, which can hurt our model performance.

\section{Conclusion}
In this paper, we investigate a novel task named Fine-grained Category Discovery under Coarse-grained supervision (FCDC), which can reduce significant labeling cost and adapt models to novel categories of different granularity from known ones. We further propose a hierarchical weighted self-contrastive model to approach the FCDC task by better controlling intra-class and inter-class distance. 
By performing supervised and contrastive learning on shallow and deep layers of pre-trained models, our model can learn fine-grained knowledge from shallow to deep with only coarse-grained supervision. Extensive experiments on public datasets show that our approach is more effective and efficient than compared methods.

\section*{Limitations}
The limitations of our method lies in two aspects. Firstly, following previous works, we need to know the number of fine-grained clusters \textit{K} as prior knowledge, which is usually difficult to get in real-world scenarios. Secondly, our method cannot predict semantic meanings (e.g., label names) of discovered fine-grained categories, which is also an unexplored question in the field of novel category discovery.

\section*{Acknowledgements}
This work was supported by National Key Research and Development Program of China (2020AAA0108800), National Natural Science Foundation of China (62137002, 61721002, 61937001, 61877048, 62177038, 62277042). Innovation Research Team of Ministry of Education (IRT\_17R86), Project of China Knowledge Centre for Engineering Science and Technology. MoE-CMCC ``Artifical Intelligence''  Project (MCM20190701), Project of Chinese academy of engineering ``The Online and Offline Mixed Educational ServiceSystem for 'The Belt and Road’ Training in MOOC China''. ``LENOVO-XJTU'' Intelligent Industry Joint Laboratory Project.

% Entries for the entire Anthology, followed by custom entries
\bibliography{custom}
\bibliographystyle{acl_natbib}

\end{document}